\documentclass{article}

\usepackage{PRIMEarxiv}

\usepackage[utf8]{inputenc} 
\usepackage[T1]{fontenc}    
\usepackage{hyperref}       
\usepackage{url}            
\usepackage{booktabs}       
\usepackage{amsfonts}       
\usepackage{nicefrac}       
\usepackage{microtype}      
\usepackage{lipsum}
\usepackage{fancyhdr}       
\usepackage{graphicx}       
\graphicspath{{media/}}     

\title{Improving Fairness in Credit Lending Models using Subgroup Threshold Optimization
}

\author{
  Cecilia Ying \\
  Department of Management Analytics\\
  Smith School of Business, Queen's University\\
  Kingston, Ontario, Canada\\
  \texttt{cecilia.ying@queensu.ca} \\
   \And
 Stephen W. Thomas \\
  Department of Management Analytics\\
  Smith School of Business, Queen's University\\
  Kingston, Ontario, Canada\\
  \texttt{stephen.thomas@queensu.ca} \\
}

\begin{document}
\maketitle

\begin{abstract}
In an effort to improve the accuracy of credit lending decisions, many financial intuitions are now using predictions from machine learning models. While such predictions enjoy many advantages, recent research has shown that the predictions have the potential to be biased and unfair towards certain subgroups of the population. To combat this, several techniques have been introduced to help remove the bias and improve the overall fairness of the predictions. We introduce a new fairness technique, called \textit{Subgroup Threshold Optimizer} (\textit{STO}), that does not require any alternations to the input training data nor does it require any changes to the underlying machine learning algorithm, and thus can be used with any existing machine learning pipeline. STO works by optimizing the classification thresholds for individual subgroups in order to minimize the overall discrimination score between them. Our experiments on a real-world credit lending dataset show that STO can reduce gender discrimination by over 90\%.
\end{abstract}


\section{Introduction}

Total Canadian non-mortgage loans have risen to \$1.7 trillion in Q4 of 2020~\cite{bank_of_canada_chartered_2021}. Lending institutions have an increasing need to develop automated credit assessment methodologies that are effective and efficient, while also adhering to national and international regulatory requirements. The recent development of machine learning (ML) techniques for default prediction has enabled innovative ways for  automated credit scoring and, in many cases, credit approvals.

However, ML-based techniques are not always perfect. In 2019, Apple’s new credit card service, \textit{Apple Card}, allegedly discriminated against female applicants, leading to angry customers, negative publicity, and a formal investigation by the New York’s Department of Financial Services \cite{vigdor_apple_2019}. Another e-commerce giant, Amazon, had to discontinue the use of its AI recruitment tool after discovering that it discriminates against female applicants as a result of the male-dominated training data used to build their model \cite{dastin_amazon_2018}.  Similar biases have also been found in language models used for machine translation tasks, in which the representation of words often reproduce stereotypes and biased associations, such as ``Man is to Computer Programmer as Woman is to Homemaker"~\cite{bolukbasi_man_2016}.

Examples like these rightfully raise ethical concerns and drive the urgent need to address biases in ML algorithms~\cite{suresh_framework_2020}. Indeed, policymakers around the world are racing to regulate the use of sensitive data including information on gender, race, religion, national origin, age, and sexual orientation~\cite{GDPR_2016}.  Policymakers have created anti-discriminatory laws with the intent to promote the ethical use of ML models, and to combat discrimination towards any protected or minority groups \cite{ellis2012eu, wachter2017counterfactual}. For example, in the United States, the Equal Credit Opportunity Act (ECOA) restricts credit lenders from collecting and using gender information in their lending decisions~\cite{knight_ai_2019}. 

However, a recent study found that, counter-intuitively, creditworthy female borrowers suffer an even \textit{higher} loan rejection rate when gender information is excluded as a feature from the prediction model \cite{kelley_anti_2020}. That is, by adhering to the ECOA, an ML model would actually be \textit{more unfair} to female borrowers.  

In this paper, we propose a solution to the unfavorable situation when gender information is excluded from the model, resulting in a higher false loan rejection rate for female borrowers . Through an experiment on the Home Credit dataset \cite{home_credit_group}, we find that creditworthy females are denied loans between 1--6\% more often when gender is excluded from the model, leading to an unfair lending situation. We postulate that the main reason for the unfairness is largely due to \textit{historical bias} present in the training data.  These creditworthy females were essentially miscategorized as ``less creditworthy males''. We propose a new technique, called \textit{Subgroup Threshold Optimization (STO)}, that is able to reduce the impact of historical bias and therefore improve fairness by up to 98\%.

The rest of this paper is organized as follows. Section \ref{sec:background} outlines existing efforts to quantify and address bias in ML models. Section \ref{sec:sto} motivates and introduces our proposed technique, STO. Section \ref{sec:experiment} describes our experiment and results. Section \ref{sec:discussion} discusses the STO and outlines future research avenues, and Section \ref{sec:conclusion} concludes.

\section{Background and Related Work}\label{sec:background}

We now briefly describe existing research to identify and correct for bias in ML pipelines. For the purposes of this work, we define \textit{bias} to be ``prejudice in favor of or against one thing, person, or group compared with another, usually in a way considered to be unfair'', and resulting in a disparity between individuals or groups that can be measured by the differences in the classification error between them \cite{corbett2018measure}. While there dozens of types of human biases exist, such as unconscious bias, confirmation bias, self-serving bias, and framing, all of which could be introduced into the ML pipeline in various stages, we will highlight below the main culprits in ML. No matter the type of bias or the ML stage, bias in the ML pipeline will lead to unfair predictions.

\subsection{Biases in Classification Models} 

\begin{figure}[t]
    \centering
    \includegraphics[width=\textwidth]{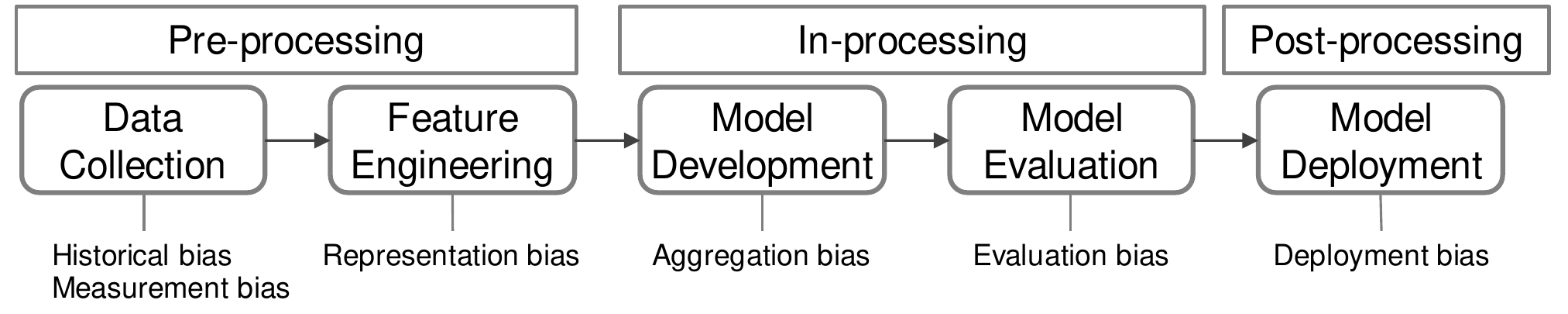} 
    \caption{Biases in a the different stages of an ML Pipeline. The typical ML pipeline consists of three main stages (i.e., pre-processing, in-processing, and post-processing). Different types of biases can be introduced at each stage. Researchers have proposed fairness solutions that work at any of the three stages.}
    \label{fig:biasPipeline}
\end{figure}

Different types of biases can be introduced --- intentionally or unintentionally --- through one or multiple stages in the ML pipeline \cite{suresh_framework_2020}, as illustrated in Figure \ref{fig:biasPipeline}. 

First, biases could be introduced during the preprocessing phase, in which training data is collected, sampled, and preprocessed. In particular, \textit{historical bias} is a result of changes of values or beliefs in the real world that have not been reflected properly in the data. Additionally, \textit{measurement bias} can occur when unsuitable proxy data is used to represent the real observable or unavailable data. \textit{Representation bias} may arise when data sample is constructed in such a way that it fails to represent the population in a general way.

Second, biases could be introduced during the in-processing phase.
The classification model itself may also generate \textit{aggregation bias} during the modelling phase, when unique features representing distinct populations are lost during aggregation.  \textit{Evaluation bias} can occur if the testing data or selected performance metric were ineffective in choosing the best model, resulting in a model that is trained and rewarded to produce biased outcomes.

Finally, biases could be introduced during the post-processing phase. Once a model is deployed, systems that update using real-time data that were previously unseen during training may introduce new bias, also known as \textit{deployment bias}.  This could lead to erroneous results generated from an unstable and unreliable system. 

\subsection{Measuring Fairness: Individuals, Groups, and Subgroups}

Existing research most commonly defines and measures fairness at either the \textit{individual} level or at the \textit{group} level, where groups are defined as a collection of individuals differentiated by a single attribute, such as gender~\cite{mehrabi_survey_2021}. Briefly, a model is considered to be \textit{fair} if it produces comparable utility measures for individuals or groups with similar attributes and if it produces equalized odds for an accurate prediction among all individuals or groups \cite{dwork_fairness_2012, hardt_equality_2016, saxena_how_2020}.  

However, there are practical challenges with the definitions on individual and group fairness. For example, at the individual level, it is often not possible to measure all aspect of all individuals, especially in a small sample or when obtaining the information is expensive and laborious \cite{mehrabi_survey_2021}. Any similarity comparison between individuals is therefore imprecise, making it difficult to perfectly assess fairness between those individuals. Group measures, however, cannot properly represent all the nuances between individuals within the group. For example, the model outcomes might appear to be fair when measured between males verse females but are unfavorable for minority females due to a mismatch in utility between disadvantaged groups and the population \cite{ben-porat_protecting_2021}.  

A third class of fairness measurements work at the  \textit{subgroup} level, which accounts for multiple individual attributes at once. Subgroup measurements has been shown to provide a better overall picture of the fairness of a model~\cite{kearns_preventing_2018}.

\subsection{Debiasing Techniques} 

Techniques for addressing the aforementioned biases can be categorized in three main categories: pre-processing, in-processing and post-processing, based on which stage of the ML pipeline they are applied, as shown in Figure~\ref{fig:biasPipeline}.

\textit{Pre-processing} techniques focus on altering the training data in some way prior to the data being given to the  ML algorithm \cite{calmon_optimized_2017, kamiran_data_2012}. The main goals of these techniques are to detect and remove historical, representation,  or measurement bias. 

\textit{In-processing} techniques attempt to adjust the machine learning algorithm and/or the evaluation steps within model development in an effort to reduce aggregation bias and evaluation bias~\cite{woodworth_learning_2017, feldman_designing_2020}. 

Finally, \textit{post-processing} techniques adjust the model's probability estimates in attempt to correct for biases that occur anywhere within the ML pipeline \cite{kim_fairness_2018, kearns_preventing_2018}.

\section{Subgroup Threshold Optimization}\label{sec:sto}

We now describe \textit{STO}, a post-processing technique to improve fairness in ML predictions. We first define a common utility measure of assessing the predictive performance of an ML model, and show how to use that measure to assess the predictive performance on subgroups of the data. We then define a \textit{discrimination score} to quantify the difference of a model's predictive performance across different subgroups. Finally, we outline the steps that $STO$ uses to minimize the discrimination metric and therefore optimize subgroup fairness.

\subsection{Thresholds, Utility Measures, and Subgroups}

\begin{figure}[t] \label{fig: STO-1}
     \centering
     \includegraphics[width=0.6\textwidth]{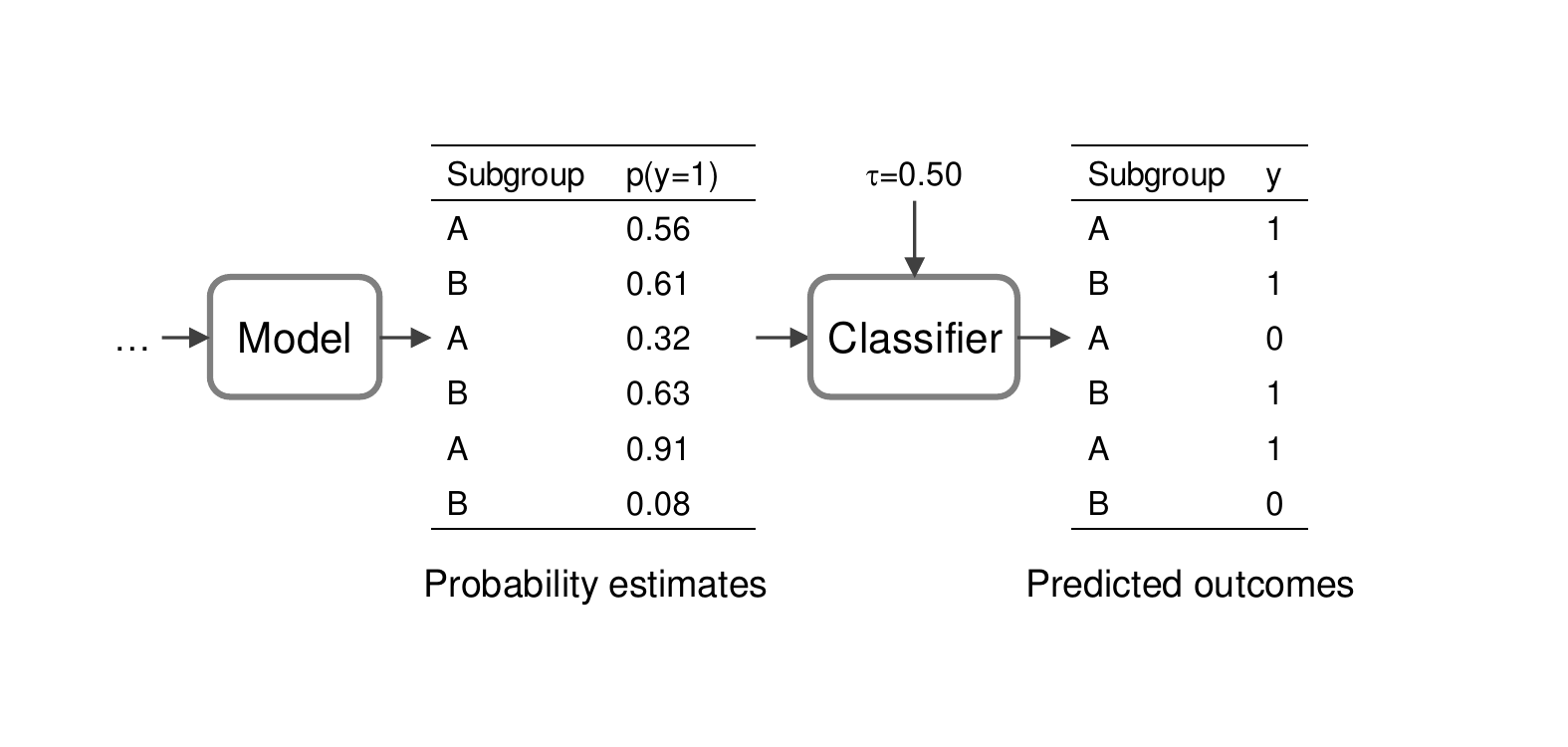} 
     \caption{An ML model predicts probability estimates for all data instances, and a single threshold $\tau$ is used to classify each instance into outcomes.}
\end{figure}

Given a data instance $x$ of a dataset $X$, an ML classification model will output a \textit{probability estimate} $p(y=1) \in [0,1]$, the probability that $x$ belongs to class $y$. The actual class membership (or\textit{ predicted outcome}) of instance \textit{x} is then determined by comparing its probability estimate $p(y=1)$ against a pre-determined threshold $\tau \in [0,1]$, as shown in Figure~\ref{fig: STO-1}. In a binary classification task, the prediction outcome is said to be  \textit{positive} (or, for example, \textit{defaulted} in a credit lending model) if $p(y=1) \geq \tau$, \textit{negative} (for example, \textit{paid}) otherwise.

Given these predictions, many utility measures exist to quantify the predictive performance of a model. One of the most popular utility measures in the credit lending domain is \textit{positive predictive value} (\textit{PPV}), also known as \textit{precision}. To calculate \textit{PPV}, we first use a historical dataset to mark each prediction as either a true positive ($TP$), true negative ($TN$), false positive ($FP$), or false positive ($FP$). \textit{PPV}  then calculates the ratio of  true positives to the total number of positive predictions (that is, true positives and false positives):

\begin{equation} \label{eq:PPV}
PPV = \frac{TP}{(TP+FP)}  
\end{equation}

Note that the number of true positive predictions and false positive predictions of a model is a function of $\tau$ and therefore changes according to the selected value of $\tau$.  For a given value of $\tau$, we call the number of true positive predictions $TP_\tau$ and the number of false positive predictions $FP_\tau$.

\begin{figure}[t]\label{fig: STO-2}
     \centering
     \includegraphics[width=0.6\textwidth]{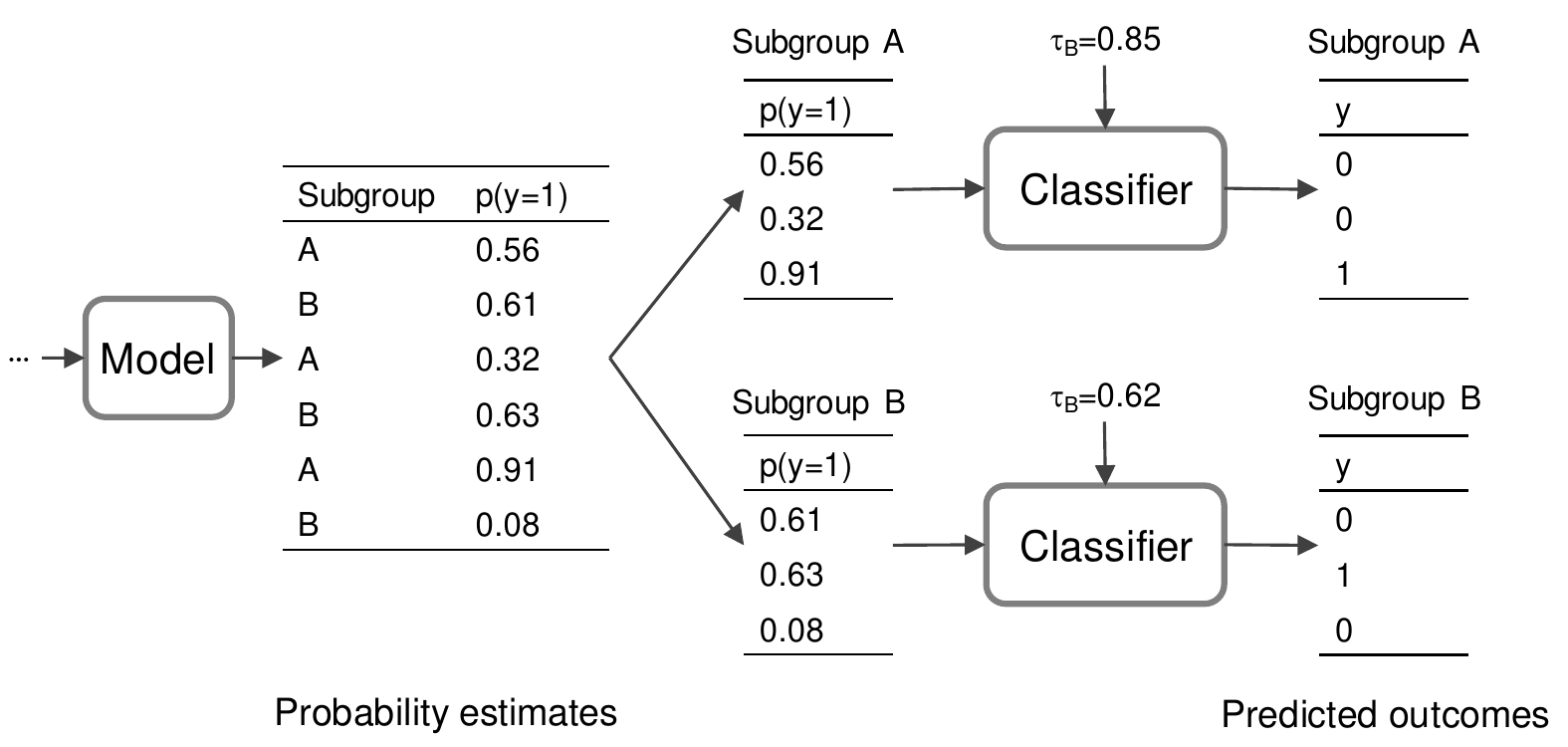} 
     \caption{An ML model predicts probability estimates for all data instances. The estimates are split into subgroups and a separate threshold is used to classify each subgroups' instances into outcomes.}
     \label{fig:STO-b}
\end{figure}

Normally, \textit{PPV} is assessed by using all data instances in $X$. However, it is possible to first group related data instances together into some number of disjoint subgroups, and calculate \textit{PPV} separately for each subgroup, as shown in Figure~\ref{fig: STO-2}. For instance, we might group all female applicants into one subgroup and all male applicants into another subgroup, and calculate \textit{PPV} separately for each. Doing so would give us an indication as to how well the model performs for each subgroup. If the subgroup \textit{PPV} values are different, then the model will lead to worse outcomes for the subgroup with lower scores.

To help keep track of the \textit{PPV} for each value of $\tau$ and each subgroup, we define the positive predictive value, $PPV_{\tau,i}$, for threshold $\tau$ and subgroup \textit{i} as:
\begin{equation} \label{eq1}
  PPV_{\tau,i}=\frac{TP_{\tau,i}}{(TP_{\tau,i}+FP_{\tau,i})}  
\end{equation}

\subsection{Discrimination Score}

We now define a score to compare a model's performance on one subgroup with that of another subgroup. Using  previous research as inspiration~\cite{kelley_anti_2020}, we define the \textit{discrimination score}, $D_{\tau_i, \tau_j}$, between subgroup \textit{i} and \textit{j} as the difference between the \textit{PPV} between the two subgroups as follows:

\begin{equation}  \label{eq:discrimination}
    D_{\tau_i, \tau_j}=PPV_{\tau,j}-PPV_{\tau,i}= \frac{TP_{\tau,j}}{(TP_{\tau,j}+FP_{\tau,j})}-\frac{TP_{\tau,i}}{(TP_{\tau,i}+FP_{\tau,i})}    
\end{equation}

In this way, we can quantify whether and how much a model discriminates against a particular subgroup. For example, we might create subgroups based on gender, and we might find that a model has a \textit{PPV} of 80\% for the male subgroup but only 70\% for the female subgroup. Such a model would have a discrimination score of 10\% against the female subgroup, as it would wrongly deny loans to more creditworthy females than it would to creditworthy males.

If a subgroup \textit{k} has a higher \textit{PPV} than all other subgroups \textit{i} within the dataset \textit{X}, we say it is \textit{strictly dominating} as it has a (possibly unfair) absolute advantage over all other subgroups.

\begin{equation}\label{eq:dominating}
   PPV_{\tau,k} \geq PPV_{\tau,i} \hspace{3mm} \forall  \hspace{1mm} i \in X 
\end{equation}

\subsection{Subgroup Threshold Optimizer (STO)}

We formulate STO as an optimization technique that takes as input a dataset $X$, a partitioning of $X$ into any number of disjoint subgroups, a model $M$, and a baseline threshold $\tau_{base}$. STO uses the following steps to find a threshold for each subgroup such that the the discrimination score between subgroups is minimized.

\textbf{Step 1: Calculate baselines.} STO uses $\tau_{base}$ to calculate the baseline \textit{PPV} for each subgroup $i \in X$ using Equation~\ref{eq1}. Similarly, STO calculates the  baseline total \textit{PPV} for all data instances. STO can then identify the dominating subgroup \textit{k} as defined by Equation~\ref{eq:dominating}.

\textbf{Step 2: Optimize.} STO finds thresholds $\tau_i^*$ for all non-dominating subgroups, as per Equation \ref{eq4}, to minimize the discrimination difference with the identified dominating subgroup \textit{k}.

\textbf{Step 3: Validate.} STO ensures that the identified optimal threshold $\tau_i^*$ for all $i$ meets the conditions stated in Equation \ref{eq5} and \ref{eq6}, such that the adjusted \textit{PPV} for individual subgroups and total aggregate \textit{PPV} at least equal to the baseline \textit{PPV}s in Step 1.

In particular, the optimization objective of Step 2 is to minimize the discrimination score, $D_{\tau_k, \tau_i}$, by reducing the difference in \textit{PPV} between the discriminated subgroups \textit{i} and the dominating subgroup \textit{k}. 

\begin{equation}  \label{eq4}
    \min D_{\tau_k, \tau_i} = \min(PPV_{\tau,k} - PPV_{\tau, i}) \hspace{3mm} \forall \hspace{1mm} i \in X
\end{equation}

STO achieves this by solving for an optimal threshold unique for each subgroup. Such thresholds will improve the \textit{PPV} for each discriminated subgroup, while at least maintaining, but possibly improving, the $PPV$ for the dominating subgroup, such that:
\begin{equation} \label{eq5}
    PPV_{\tau_{base}, k} \geq PPV_{\tau_{adj}, i} \geq PPV_{\tau_{base}, i} \hspace{3mm} \forall \hspace{1mm} i \in X 
\end{equation}

Additionally, by the additive property, we see that the $PPV$ for the entire group will not decrease:
\filbreak
\begin{equation} \label{eq6}
    PPV_{\tau_{adj}, all} \geq PPV_{\tau_{base}, all}
\end{equation}

\section{Experimental Results}\label{sec:experiment}

As mentioned in Section~\ref{sec:background}, fairness can be defined at either the group level or the subgroup level. To quantify the benefits of STO in both situations, we performed two experiments, both in the credit lending domain. In the first experiment, we use the gender attribute to create groups of individuals. In the second experiment, we use an unsupervised ML techniques which combines many individual attributes to create subgroups. In both experiments, we calculate the discrimination score between subgroups both before and after STO is used. 

Note that for terminology ease, we refer to both groups and subgroups as ``subgroups'' in the following descriptions.

\subsection{Experiment Setup}
The data used for both experiments is provided by Home Credit Group, an international consumer finance provider founded in 1997 currently operating in nine countries~\cite{home_credit_group}. The data includes borrowers' information such as history of credit card balances, previous loans and repayment history, and various sociodemographic information, including gender, for a total of 134,158 instances and 219 features.  For our experiment, we examine a subset of the 122 features available in the main data file with key information about the loan and the borrower at the time of their application. The data can be obtained online~\cite{kagglecom_home}.

We experimented with three standard, popular classification algorithms: Decision Tree, Logistic Regression, and Random Forests. 


To understand the role of gender in this dataset, we compared the \textit{PPV}  between male and female applicants across all three models. The results are shown in Table \ref{tab:PPV_perModel}.  We found that all three models consistently under performed for females, with a lower PPV resulting from a higher number of false positive predictions,  leading to wrongfully rejecting creditworthy females.

\begin{table} 
 \caption{PPV Summary (per Model per Gender)}
  \centering
  \begin{tabular}{llll}
    \toprule
    Model 
        & $PPV_{all}$& $PPV_{male}$&$PPV_{female}$ \\
    \midrule
    Decision Tree 
        & 0.8346 & 0.8880 & 0.7483 \\
    Logistic Regression
        &  0.5467 & 0.6964 & 0.3724\\
    Random Forest
        & 0.7339 & 0.8197 & 0.6112 \\
    \bottomrule
  \end{tabular}
  \label{tab:PPV_perModel}
\end{table}

The Decision Tree model had the best performance among the three models, with an \textit{PPV} score of of 0.8346. Therefore, for our experiments probability estimates were generated from the Decision Tree model. 

\subsection{Experiment 1: Gender Groups}
In this experiment, we test the impact of STO on group fairness. We defined the groups using on a single predetermined protected attribute, gender. In this case, as gender was a binary attribute, there are two subgroups: male and female.

We set the baseline threshold $\tau = 0.50$. STO calculated the aggregated \textit{PPV} of the entire dataset, the \textit{PPV} for the male subgroup, and the \textit{PPV} for the female subgroup. STO then calculated the discrimination score as defined by Equation~\ref{eq:discrimination}. Table \ref{tab:exp1} shows the results.  

With the baseline threshold, we found that the \textit{PPV} for the male subgroup, 0.7713, is higher than for the female subgroup, 0.6836, leading to a discrimination score of 0.0877. The male subgroup is therefore the dominating subgroup as defined by Equation~\ref{eq:dominating}.

Next, STO found optimal thresholds for the two subgroups. While keeping $\tau_{male}$ constant, STO searched for an optimal threshold $\tau_i^*$ for the female subgroup that minimizes the discrimination score $D$. STO found the optimal threshold for the female subgroup to be $\tau_{adj, female}^*=0.65$, leading to an improved $PPV_{\tau,{female}}$ of 0.7639. The discrimination score was reduced to 0.0073, a reduction of 91.7\%. Furthermore, the   PPV of the dominating class was not decreased, and the overall aggregated \textit{PPV} was improved by 7.9\% and \textit{PPV} for the non-dominating class by 117.5\%.

\begin{table} 
 \caption{Experiment 1 results. STO is able to increase the PPV of the female subgroup from 0.6836 to 0.7639 (+117.5\%) without decreasing the PPV of the male subgroup. The discrimination score of the female subgroup decreases by 91.7\%.}
  \centering
  \begin{tabular}{llllll}
    \toprule
   & $\tau$  & $PPV_{\tau, all}$& $PPV_{\tau,{male}}$ & $PPV_{\tau,{female}}$ & $D_{\tau_{male}, \tau_{female}}$ \\
    \midrule
    Baseline & 0.50 
        & 0.7107 & \textbf{0.7713} & 0.6836 & 0.0877 \\
    $\tau_{adj, female}^*$ & 0.65
        & 0.7669 & 0.7713 & 0.7639 & 0.0073 \\
    \midrule
    Net Diff. &
        & +0.0562 & - & +0.0803 & -0.0804 \\
    \% Diff. & 
        & +7.9\% & - & +117.5\% & -91.7\% \\
    \bottomrule
  \end{tabular}
  \label{tab:exp1}
\end{table}

\subsection{Experiment 2: Cluster Subgroups}

We next test the impact of STO on subgroup fairness. In order to test the generality of STO, we defined each subgroup not by an individual protected attribute in the data, but instead using an unsupervised clustering technique. We determined the number of clusters \textit{k} by using the elbow plot method, optimizing inertia vs. $k$ \cite{sinaga2020unsupervised}.   In this dataset, we found the optimal $k$ to be 4.

Experiment 2 is different from Experiment 1 in two ways. First, by assigning a cluster label through an unsupervised learning technique, we are not restricting the subgroup definition to be based on a single predetermined protected attribute and therefore will be able to capture additional interactions between the features to better identify the similarity between each instance. Second, the search for the optimal set of thresholds $\tau_i^*$ now becomes a multi-dimensional optimization problem to minimize the discrimination score $D_{\tau_i, \tau_j}$ for each subgroup pair. However, we show that this is still a bounded problem that can be efficiently solved, restricted by the condition where the dominating subgroups must maintain or improve its subgroup utility post adjustments.

We created the subgroups using the cluster IDs and we set the baseline threshold $\tau = 0.50$. STO then evaluated the baseline \textit{PPV}s as shown in Table \ref{tab:exp2}. We found the entire dataset has a \textit{PPV} of 0.7109, while the \textit{PPV} of the subgroups range from 0.6752 (for subgroup $C0$) to 0.7597 (for subgroup $C2$). Subgroup $C2$ therefore is the dominating subgroup while subgroup $C0$ has the highest discrimination score of 0.0845.


STO used the \textit{PPV} of the dominating subgroup, $C2$, as the benchmark to evaluate the updated \textit{PPV}s from the adjusted thresholds for each non-dominating subgroup. STO adjusted the discrimination scores, $D_{C2,Ci}$,  by calculating the difference between the \textit{PPV} of subgroup $C2$ and the \textit{PPV} for subgroups $C0$, $C1$, and $C3$. STO searched for the set of thresholds that minimize the discrimination scores for each pair.

STO found the optimal thresholds for each of the four subgroup to be 0.65, 0.60, 0.55, 0.60, respectively. We compared the PPVs against the baseline threshold and found that the overall \textit{PPV}s for the entire dataset increased, from 0.7109 to 0.7649, providing an overall dataset-level improvement of 7.6\% in accuracy.  The \textit{PPVs} also improved for all four subgroups with the new thresholds, from 1.1\% to a 13.4\%. Discrimination scores were also reduced by 82.1\%--97.9\%.

\begin{table} 
 \caption{Results for Experiment 2. Using the baseline threshold of 0.50, we find that the \textit{PPV} scores of the four subgroups vary from 0.6752 to 0.7597. STO finds optimal thresholds that increase \textit{PPV} by 1.1\%--13.4\% and reduce discrimination scores by up to 97.9\%.}
  \centering
  \footnotesize
  \begin{tabular}{lllllll | llll}
    \toprule
   & $\tau$  & $PPV_{\tau, all}$
   & $PPV_{\tau,{C0}}$ & $PPV_{\tau,{C1}}$ 
   & $PPV_{\tau,{C2}}$ & $PPV_{\tau,{C3}}$ 
   & $D_{C2,Ci}$ & $D_{C2,Ci}^*$ & Diff. & \% Diff.\\
    \midrule
    Baseline & 0.50  & 0.7109 
        & 0.6752 & 0.7173 
        & 0.7597 & 0.7256 &\\
    \midrule
    $\tau_{adj,C0}^*$ & 0.65  &  
        & 0.7659 &  
        &  &  
        &0.0845 & 0.0017& -0.0827& -97.9\%\\
    $\tau_{adj,C1}^*$ & 0.60  &  
        &  & 0.7601
        &  &  
        &0.0424 & 0.0071& -0.0348& -82.1\%\\
    $\tau_{adj,C2}^*$ & 0.55  &  
        &  &  
        & 0.7677 &  
        &- & -&&\\
    $\tau_{adj,C3}^*$ & 0.60  &  
        &  &  
        &  & 0.7660 
        &0.0341 &0.0016& -0.0324& -95.0\%\\
    \midrule
    Adj. PPV & &0.7649 
    & 0.7659 & 0.7601
    & 0.7677 & 0.7660
    & \\
    \midrule
    Diff. & & +0.0540
    & +0.0907 & +0.0428
    & +0.0080 & +0.0404
    & \\
    \% Diff. & & +7.6\%
    & +13.4\% & +6.0\%
    & +1.1\% & +5.6\%
    & \\
    \bottomrule
  \end{tabular}
  \label{tab:exp2}
\end{table}

\section{Discussion and Future Work}\label{sec:discussion}

Even though there has been an expansion of research related to fairness in ML, the main body of work remains mostly theoretical. The different definitions of fairness and different measures of discrimination continue to be the main discussion points for researchers, in a quest to find a unified and systematic way to combat hidden biases in predictive models \cite{chouldechova_snapshot_2020}. In this paper, we took a practical approach to put theory into action for assessing prediction models in the credit lending domain, with the primary goal of creating a flexible technique that can easily be applied by non-experts in the credit lending industry.

Our current methodology is a post-processing technique that can use the probability estimates from any classification model to debias the outcome across any subgroups based on a user-defined baseline threshold and any utility measure to evaluate the proposed updates. This approach provides a simple and interpretable approach for end users without any machine learning expertise to deploy into their existing prediction model pipeline and integrate a human-in-the-loop operation.

Our experiments used Decision Trees as the main classification algorithm, but STO can be applied as a post-processing adjustment to any classification models, including higher-performing black-box algorithms such as Neural Networks, Support Vector Machines, and gradient boosted ensembles. STO can even be applied when the user has no access to the training data or prediction model itself.

STO is also agnostic to the process of creating the subgroups. Our experiments used protected attributes and unsupervised clustering, but any other definition would work equally well. This flexibility helps avoid, for example, the troublesome binary definition of gender, which is no longer representative of modern interpretation of gender as a non-binary fluid concept.

Risk averse credit lenders might be concerned about a guaranteed revenue from accurately predicting loans that will be paid, and avoiding loans that are predicted to be paid but would later default. STO allows the user to set the base threshold to achieve any level of risk tolerance.

Another key feature of STO is the human-in-the-loop operation. Users of STO can define the initial baseline threshold in accordance with their risk tolerance and also select the utility measure that is suitable for their use case.

Finally, STO is easy to understand, flexible to be deployed as a post-processing step, and does not require user expertise in machine learning to implement. It also provides the ability to debias the predicted outcome across all subgroups based on user defined parameters, while allowing firms to regain control over black-box models that are hard to interpret. There are also a number of model extensions that can be explored to establish the generalizability of this method in other classification tasks, and to further contribute to the literature of fairness in machine learning applications.

In future work, we aim to enhance the flexibility of STO even further by relaxing the constraint that the \textit{PPV} of the dominating subgroup must not decrease. (That is, we hypothesize that even greater fairness might be achieved in some situations if we decreased the \textit{PPV} of the dominating subgroup, which might allow us to increase the \textit{PPV} of the other subgroups even further.) 

We also plan to build a financial model that can assess how much money a firm stands to gain or lose by ensuring fairness across predictions. 

We would also like to extend STO so that it is able generate a bias-neutral synthetic dataset that remains representative of the general population of the data, and incorporate methods that requires input from human expert to preserve a human-in-the-loop approach for explainable outcomes in a practical application. Given that the current model is trained on individuals that were granted a loan, the model may never include feature that are similar to anyone who were turned away from getting a loan in the first place. This representation bias could potentially be addressed by synthetically creating the additional data to infer the missing parts of the population.

We would also expand the current empirical work to experiment with other types of classification algorithms to establish STO's generalizability for different classification tasks. Furthermore, we wish to experiment with different level of subgroups granularity, from increasing the number of clusters to individual pair-wise fairness, to support a wider range of application needs. Finally, we believe there are many potential enhancements worth exploring for extending STO's use across different industries.

\section{Conclusion}\label{sec:conclusion}

In this paper, we proposed a new interpretable post-processing technique, \textit{Subgroup Threshold Optimizer} (\textit{STO}), to enhance fairness in ML classification predictions. STO can enhance fairness across all user-defined subgroups, given a user-defined baseline threshold and a utility measure. STO finds optimal post-prediction threshold adjustments that minimize the discrimination scores for each subgroup. The adjustments ensure that the utility measures are no worse than, and are often times much better than, the baseline levels for all subgroups and for the entire dataset as a whole.

We provided empirical results from two experiments that validate the effectiveness of STO in improving group fairness as well as subgroup fairness. In our experiments, we found that STO decreased discrimination scores in PPV by over 92\% for gender groups and by over 98\% for subgroups defined by an unsupervised ML technique.

\bibliographystyle{unsrt}  
\bibliography{references} 
\end{document}